\documentclass[conference]{IEEEtran}
\IEEEoverridecommandlockouts
\usepackage{cite}
\usepackage{amsmath,amssymb,amsfonts}
\usepackage{algorithmic}
\usepackage{graphicx}
\usepackage{textcomp}
\usepackage{xcolor}
\usepackage{booktabs}
\usepackage{tikz}
\usepackage{tcolorbox}
\usetikzlibrary{arrows.meta,positioning,calc}
\definecolor{okblue}{RGB}{0,114,178}
\definecolor{okorange}{RGB}{230,159,0}
\definecolor{okgreen}{RGB}{0,158,115}
\newcommand{\AnswertoRQ}[2]{%
  \begin{tcolorbox}[colback=black!1!white,colframe=black!40!white,
    left=0.88mm, right=0.88mm, top=0.88mm, bottom=0.88mm, boxsep=0.66mm, arc=2.5mm]
    \textbf{Answer to #1:} #2
  \end{tcolorbox}}
\usepackage{url}
\def\BibTeX{{\rm B\kern-.05em{\sc i\kern-.025em b}\kern-.08em
    T\kern-.1667em\lower.7ex\hbox{E}\kern-.125emX}}
\begin{document}
\bstctlcite{BSTcontrol}

% 务必确保导言区有这一句解锁脚注
\IEEEoverridecommandlockouts

\title{SenWorld: A Digital-Twin Simulation for Generating Context-Rich Evaluation Data}

\author{
\IEEEauthorblockN{Zenghui Zhou, Xiaoyang Li, Xiaoxuan Qiao, Zhilang Wei, and Tianming Lei\textsuperscript{*}\thanks{\textsuperscript{*}Corresponding author.}}
\IEEEauthorblockA{\textit{ByteDance Inc.}, Beijing, China\\
zhouzenghui.p@gmail.com, \{lixiaoyang.lxy1, qiaoxiaoxuan, weizhilang, leitianming\}@bytedance.com}
}

\maketitle

\begin{abstract}
Smartphone personal assistants reason over longitudinal personal data, yet evaluating them requires context-rich evaluation data whose correct answers are known, and real device traces are too privacy-sensitive to share.
To address this challenge, we present SenWorld, a physically grounded, deterministic, event-sourced digital-twin simulation that generates such data with ground truth fixed by construction. 
In SenWorld, personas live through a full day in a world built from real map, weather, holiday, and network data; every observable signal is archived in full-system snapshots; and each evaluation case is labeled by a pointer to an existing record rather than by post-hoc annotation or a large language model (LLM) judge.
We evaluate this method with 16 personas in Beijing.
The generated data closely matches the held-out real-user benchmark in category distribution (Jensen--Shannon divergence (JSD) 0.070) and in the daily rhythm of communication records (JSD below 0.1), though generated records remain shorter than real ones.
Without scripted interaction, personas form a fully reciprocated dialogue subgraph and differentiated behavioral repertoires.
Projected into 717 evaluation cases, the generated data runs end-to-end against a production smartphone assistant: the assistant answers 98.9\% of the cases, and all eight residual failures fall on call and Short Message Service (SMS) records while contacts, schedules, notes, and alarms never fail.
Each failure remains diagnosable through its snapshot pointer, with no LLM judge involved.
Overall, SenWorld offers a privacy-safe, reproducible, and distribution-checked path to evaluation data whose labels are fixed by construction.
\end{abstract}


\section{Introduction}
Smartphone personal assistants are increasingly deployed as autonomous software agents on personal devices~\cite{Li2024PersonalLLMAgentsSurvey}.
They retrieve private information, interpret user context, and execute device actions.
Operating directly on personal data and device state makes their failures qualitatively different from ordinary conversational errors: a misread context or an incorrect retrieval can surface stale data or trigger unintended device actions~\cite{Rawles2025AndroidWorld,Kong2025MobileWorld}.
As these assistants become more capable of reasoning over stateful, longitudinal personal data, pre-release evaluation becomes critical~\cite{Chang2024LLMEvaluationSurvey}.

This evaluation is particularly demanding for multi-hop requests that span multiple data sources and require longitudinal context.
For instance, when a user arrives at a friend's apartment and asks the assistant to ``call the friend who lives here,'' a correct response requires the current location, the contact whose address matches it, and that contact's number.
These records must genuinely exist and interlink in the on-device state.
Testing such requests demands context-rich personal-state data where the relevant records, temporal relations, and expected answers are definitively known.
While real smartphone traces are the most natural source of such data, they are fundamentally encumbered by privacy concerns, making them unshareable and unsuitable as reproducible evaluation assets.
Consequently, the evaluation challenge is inherently a data-production problem: how can we synthesize realistic personal-state data that is privacy-safe, reproducible, and paired with reliable ground truth?

Existing data-generation methods struggle to satisfy these requirements simultaneously.
Manually constructed cases are safe and controllable but fail to capture the longitudinal complexity and long-tail combinations of real device states.
Generation with large language models (LLMs) can scale volume, but typically lacks an underlying device state, leaving correctness verification to fragile manual checks or LLM judges~\cite{Gu2026LLMasaJudgeSurvey,Molina2025TestOracleAutomationLLMs}.
While recent state-grounded methods~\cite{Khedar2026StateGen} fix truth over an underlying state, that state remains an abstract task backend rather than the actual smartphone ecosystem an assistant must navigate, and their correctness scoring still passes through an LLM judge.

Our key insight is that simulated device state can serve as evaluation data whose labels are fixed by construction for objectively world-determined properties, provided the simulation fully owns that state and evolves it strictly under real-world physical constraints.
At any simulated moment, the correct answer to a query is not inferred post-hoc by a model.
It is intrinsically determined by the exact records existing in the simulated smartphone state.
This changes the question: not whether a generated answer is plausible, but whether the generated state is realistic, reproducible, and rich enough to support testing.

We operationalize this insight with SenWorld, a physically grounded, deterministic, event-sourced digital-twin simulation.
SenWorld advances personas, smartphones, external services, and physical contexts as a single, cohesive evolving world, exporting a full-system snapshot at every simulated frame.
For each evaluation query, the ground-truth answer is simply encoded as a definitive pointer into the snapshot (a source-table identifier and a unique record ID).
Evaluating a model thus reduces to verifying whether the agent locates the correct target record.
While natural-language queries may be synthesized by an LLM, the ground-truth label strictly bypasses any LLM judge.

We evaluate SenWorld along three dimensions: we compare the distribution of the generated data with a held-out real-user benchmark across six shared categories, examine what behavior and interaction data emerges across personas without hand-written interaction scripts, and ask whether the generated evaluation cases expose and help diagnose defects in a production smartphone assistant.

This paper makes three contributions:
\begin{itemize}
\item A physically grounded, deterministic digital-twin simulation for producing context-rich evaluation data, offering ground truth fixed by construction without any LLM judge.
\item A snapshot-centered projection mechanism that deterministically translates one simulated world state into multiple downstream testing artifacts.
\item Three empirical evaluations demonstrating distribution-level fidelity to a held-out real-user benchmark, autonomous emergence of behavioral patterns, and end-to-end utility for judge-free evaluation of a production assistant.
\end{itemize}

The rest of this paper is organized as follows: Section II gives the background, and Section III presents the SenWorld method.
Section IV describes the evaluation setup, and Section V reports the results.
Section VI concludes.

\section{Background}
Modern smartphone assistants are shifting from rule-based command tools to LLM agents that reason on users' personal data.
Evaluating these agents requires datasets covering complete daily user-device interactions.
This section introduces three foundations of our framework: LLM-based smartphone assistant agents as the systems under test, generative agent simulation as a method to synthesize realistic human behavioral trajectories, and the digital twin as the provider of persistent, queryable smartphone-related state.

\textbf{Smartphone assistant agents.}
Early intelligent personal assistants rely on intent recognition and handcrafted rules, which limits their practicality and scalability~\cite{Li2024PersonalLLMAgentsSurvey}.
LLMs enable a new class of system.
Li et al.~define \textit{personal LLM agents} as LLM-based agents deeply integrated with personal data and personal devices, used for personal assistance~\cite{Li2024PersonalLLMAgentsSurvey}.
Compared with general LLM agents, they are more deeply engaged with personal data and mobile devices, and are designed to assist people rather than replace them.
Their architecture rests on three capabilities.
Task execution translates user commands into actions on personal resources.
Context sensing perceives the user's status and environment.
Memorization obtains, manages, and utilizes the user's accumulating records~\cite{Li2024PersonalLLMAgentsSurvey}.

AutoDroid, for instance, couples LLM reasoning with a structured memory of app states to automate tasks across arbitrary Android applications~\cite{Wen2024AutoDroid}.
GPTVoiceTasker uses an LLM to parse voice commands and maintains an execution memory that streamlines subsequent similar tasks~\cite{Vu2024GPTVoiceTasker}.
A vision-language model can even operate a phone from screenshots alone, emitting human-like gestures without any API access~\cite{Dorka2024VLMSmartphoneAssistant}.
Despite their different interfaces, these systems share a common dependence: the agent must retrieve and reason over the user's personal data to act correctly.
Evaluating this capability has itself become a research concern~\cite{Chang2024LLMEvaluationSurvey}, with benchmarks such as SmartBench measuring on-device LLMs on realistic mobile tasks~\cite{Lu2025SmartBench}.
However, existing benchmarks primarily focus on isolated, short-horizon tasks.
A correct answer often requires combining several personal records and their cross-source relations with the user's current situation, such as where the user is, what time it is, and what just happened.
Testing this ability therefore requires more than isolated input--output examples.
It requires a realistic personal state that carries the records, their relations, and the user's situation over a full day.

\textbf{Generative agent simulation.}
LLM-based agent simulation has become an active research area for synthesizing believable human behavior~\cite{Gao2024LLMAgentSimulationSurvey}.
Park et al.~introduced Generative Agents, which populated a small virtual town with 25 LLM-driven agents that formed social routines, maintained consistent personalities, and reacted plausibly to events over simulated days~\cite{Park2023GenerativeAgents}.
Their agents drew on persistent memory, reflection, and planning to sustain coherent behavior.
Park et al.~later scaled this idea to over a thousand agents grounded in real interviews, reproducing individual human attitudes with substantial fidelity on standard survey instruments~\cite{Park2024SimulationOfIndividuals}.
These results show that a simulated population can yield credible behavioral data at scale, which makes simulation a viable alternative to collecting real traces.

The major limitation, for our purpose, is that these simulations record what agents do and say as unstructured narrative text.
They do not maintain a device-side state that an assistant can query.
An agent does not operate on free-text traces of ``John called Mary'' but on structured app records such as a contacts database, a call log, or a notes table.
Consequently, behavioral simulation alone cannot produce the data substrate on which a smartphone assistant runs.

\textbf{Digital twins.}
A \textit{digital twin} is a simulated counterpart of a physical or software system that maintains a faithful, queryable representation of that system's state~\cite{Barricelli2019DigitalTwinSurvey}.
Digital twins have been adopted across cyber-physical and software systems as a testing substrate, where the twin stands in for the real system so that behavior can be exercised without operating the physical original~\cite{Somers2023DigitalTwinCPSTesting,GuineaCabrera2024DigitalTwinSE}.
In our context, the digital twin encompasses not just the device, but the user's entire digital and physical ecosystem.
A key property of such a digital twin is that it strictly \textit{owns} its state.
Every record in the twin is placed there by the simulation, so the twin can report exactly what that state contains.
Consequently, ground truth can be established by construction rather than relying on manual annotation or an LLM judge.
This advantage directly addresses the test-oracle difficulty in agent evaluation~\cite{Barr2015OracleProblemSurvey,Dobslaw2025LLMTestingTaxonomy}.

However, two challenges remain before this property can support assistant evaluation.
The simulated state must be realistic enough to resemble a real user's device, and its generation must be reproducible.
Meeting these two challenges, and thereby obtaining evaluation data whose ground truth is fixed by construction, is the subject of the rest of this paper.

\section{The SenWorld Method}

\subsection{Overview and Method Contract}
\begin{figure*}[t]
\centering
\includegraphics[width=\textwidth]{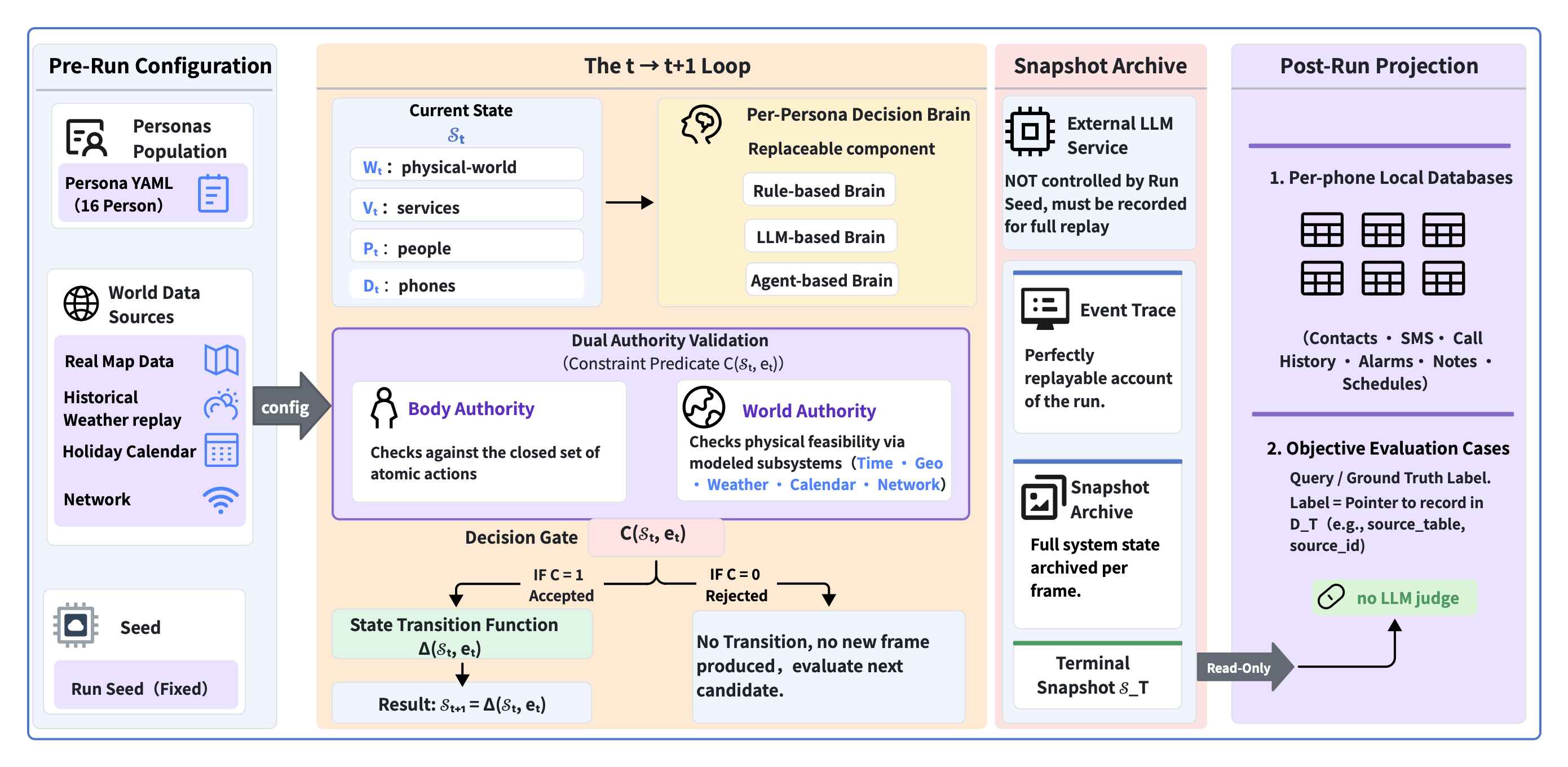}
\caption{SenWorld overview.
Personas propose actions through pluggable brains.
A deterministic, event-sourced core validates and commits every transition against a physically grounded world, persisting each event before dispatch and archiving the full system state $\mathcal{S}_t=(W,V,P,D)$ per frame.
Read-only projections then derive on-device records and evaluation cases whose labels are record pointers, not LLM judgments.}
\label{fig:core}
\end{figure*}

Figure~\ref{fig:core} shows the flow of a single run.
A run is configured by a population of personas, a physical world built from real map, weather, holiday, and network data, and a run seed.
During the simulated day, each persona repeatedly proposes its next action.
The persona's body checks the proposal against its closed set of executable atomic actions, the physically grounded world validates its physical feasibility, and every accepted action commits through the event bus as a recorded transition.
Each committed transition yields one frame, at which the full system state is archived as a snapshot.
After the run, read-only projections derive per-phone local databases and evaluation cases from the terminal snapshot without modifying the source state.
The remainder of this section formalizes this flow as a state-transition contract and then describes the mechanisms that enforce it.

SenWorld treats a day-long run as a state-transition system rather than as a text-generation procedure.
Given a persona configuration, a world configuration, and a run seed, the simulator produces an ordered event trace and a sequence of full-system snapshots.
At frame $t$, we represent a snapshot as $\mathcal{S}_t=(W_t,V_t,P_t,D_t)$, where $W_t$ is the physical-world state, $V_t$ is the service state, $P_t$ is the joint objective state of all persons, and $D_t$ is the joint state of all phones.
Let $e_t$ denote a candidate event at frame $t$, such as a proposed move or phone interaction.
A frame corresponds to a single event-triggered transition step, so multiple frames may share the same simulation time.
Let $C(\mathcal{S}_t,e_t)\in\{0,1\}$ be a binary constraint predicate that returns 1 when the event is valid in the current state and 0 otherwise.
Let $\Delta$ denote the state-transition function applied to an accepted event.
The simulator commits a transition only when the candidate event satisfies the relevant world and entity constraints:
\begin{equation*}
\mathcal{S}_{t+1}=\begin{cases}
\Delta(\mathcal{S}_t,e_t), & C(\mathcal{S}_t,e_t)=1,\\
\mathcal{S}_t, & C(\mathcal{S}_t,e_t)=0.
\end{cases}
\end{equation*}
The event trace records how the state changed, whereas the snapshot records what is true at a particular simulated frame.
A rejected candidate commits no transition and archives no frame.
The next candidate is evaluated against the unchanged state.
The downstream projection $\Pi(\mathcal{S}_T)$ reads the terminal snapshot and produces device-local records and evaluation cases without changing the source state.
For an objective evaluation case, the expected answer is the unique identifier of an existing phone state record in $D_T$, rather than a post-hoc judgment of a generated answer.

\subsection{Event-Sourced Execution}
To enforce the state-transition contract defined above, the system must record how every state change happened.
SenWorld therefore persists every candidate event $e_t$ in an append-only log before dispatching it to any subscribers.
Each event carries a stable identifier, and events scheduled for the same simulation time are consumed under a deterministic total order based on type priority and event ID.
Recording each transition before subscriber execution makes the event log a replayable account of the run.
All cross-entity state changes pass strictly through this event bus, while provider interfaces expose only read-only perceptions to the personas.
Because the core also archives a full-system snapshot $\mathcal{S}_t$ at each frame, any replay can be audited against both the historical event trace and the resulting concrete state.

Internally, all pseudorandom choices are derived from a fixed run seed, so state commits are reproducible under the ordering enforced by the event core.
However, this deterministic boundary applies exclusively to the simulator and exporter under fixed inputs.
The run seed does not control an external LLM's model version, prompt execution, sampling, retries, or service behavior~\cite{Ouyang2025ChatGPTNondeterminismCode}.
Consequently, an LLM-driven run is fully reproducible only when these external responses and execution settings are also recorded and replayed.

\subsection{Physically-Grounded State}
The method also requires that generated state be constrained by physical realities rather than merely steered by textual prompts. 
Within our state-transition contract, the physical world state $W_t$ is the authority for the physical-feasibility component of the constraint predicate $C(\mathcal{S}_t, e_t)$. 
It is governed by five modeled subsystems: time, geography, weather, calendar, and network. 
The configuration relies on a simulation clock, a graph of geocoded points of interest with a distance matrix, historical weather replay, an official holiday calendar including shifted workdays, and location-bound Wi-Fi access points with globally unique basic service set identifiers (BSSIDs). 
These subsystems determine the physical-feasibility outcome of $C(\mathcal{S}_t, e_t)$, constraining which transitions the simulator can commit and which environmental signals a phone can observe.

For example, a brain may propose a movement event $e_t$ to a new destination.
The geographic subsystem resolves the canonical point of interest and checks reachability before the virtual body is allowed to change location. 
If the destination is nonexistent or unreachable, the constraint predicate evaluates to $C(\mathcal{S}_t, e_t) = 0$.
The transition is thus strictly rejected, leaving the location state unchanged ($\mathcal{S}_{t+1} = \mathcal{S}_t$). 
Consequently, while an LLM brain can propose arbitrary behavior, it cannot overwrite or bypass modeled physical truth. 
This same authority determines objective co-presence and dynamically links device observations (such as location-bound Wi-Fi BSSIDs) to the overarching physical state. 
This keeps the simulation physically consistent by construction.
Whether the generated data resembles the held-out real-user benchmark is evaluated separately in RQ1.

\subsection{Personas, Phones, and Observable Effects}
SenWorld separates behavioral decisions, objective person state, and device-observable effects.
Each persona contains a \textit{brain} that proposes an action and a \textit{body} that checks the proposal against its closed set of executable atomic actions and executes accepted actions.
The two checks are distinct: the body determines whether the person can perform the action, whereas the world determines whether the action is physically feasible.
The brain is replaceable, for example by a rule-based or LLM-based decision component, while the body and world remain the state authorities.
A person receives world and phone observations through provider interfaces and cannot directly mutate another entity.

The phone twin is personality-free and exposes permission-level device state rather than simulator-internal state.
Its context includes battery and charging status, screen state, foreground application, Wi-Fi service set identifier (SSID) and BSSID, cellular connectivity, Bluetooth, notifications, audio session state, posture, and assistant wake-up state.
Its application stores hold records such as contacts, SMS, call history, alarms, notes, and schedules.
Persona attributes and private simulator state are not emitted as phone-side observations unless they have a corresponding device record.
When a person operates the phone, the simulator realizes the causal chain $\textit{action}\rightarrow\textit{device effect}\rightarrow\textit{phone state}\rightarrow\textit{event}$.
This separation makes user-specific behavior originate in the person while keeping the exported phone view aligned with what a real application could observe.

\subsection{Snapshot and Projection}
At each recorded frame, SenWorld serializes the system into four snapshot segments.
The physical segment contains time, locations, weather conditions, the official holiday calendar, and network facts.
The service segment contains the configured per-persona service state, such as messaging, calling, user calendar records, and weather-service content.
The person segment contains objective state and decision context.
The phone segment contains device context and the on-device databases accumulated during the run.
The snapshot is the central artifact because it retains the state from which both labels and downstream views are derived.

The service layer combines configured inputs with interactions produced inside the shared world.
For example, scenario configuration may provide an external message, while a global phone-number directory routes one persona's outgoing call or SMS to another persona's phone.

For the evaluation cases used in this paper, a read-only adapter selects an existing phone record under a predefined query template and retains its table and record identifier as the label.
The pointer refers to a record in the exported state and is not an answer invented after generation.
A checker can therefore determine whether an assistant selected the designated record without asking an LLM to judge correctness.
An LLM may assist with the natural-language surface form of a query, but it does not define the target record.
The pointer oracle supports objective properties represented in the snapshot, not subjective judgments about response quality or arbitrary open-ended explanations.

After the run, the projection reads the terminal snapshot and derives per-phone local data and evaluation cases.
Changing the persona or world configuration creates a new scenario, whereas changing only the downstream format creates another view of the same snapshot.
Under fixed input, exporter, and serialization settings, this projection is deterministic.

\subsection{Scope and Boundaries}
\label{sec:scope}
The implemented method targets multi-signal state and behavior within one smartphone per person.
The reported configuration uses Beijing physical data and a persona population designed for coverage rather than statistical representativeness.
The no-judge property is limited to objective properties represented in the snapshot and exposed by a downstream adapter.
Physical consistency does not by itself establish external behavioral realism, and some service inputs remain scripted.
An LLM brain can also produce repetitive or implausible trajectories even when the world rejects physically invalid transitions.
Earphone and watch types are architectural placeholders and do not participate in the method or evaluation reported in this paper.
We leave these limitations to future work.
They do not affect the evaluation of the method's core properties: reproducibility, physical grounding, and ground truth fixed by construction.

\section{Evaluation Setup}

\subsection{Research Questions and Study Design}
The evaluation asks three research questions (RQs) regarding the realism of the generated data (RQ1), the generation of behaviors (RQ2), and the utility of the framework (RQ3):

\begin{itemize}
    \item \textbf{RQ1 (Fidelity):} How similar is the distribution of the generated data to that of a held-out real-user benchmark?
    \item \textbf{RQ2 (Emergence):} What behavior and interaction data emerges across personas during simulation?
    \item \textbf{RQ3 (Utility):} Can the generated evaluation cases expose and help diagnose defects in a production smartphone assistant?
\end{itemize}

The three questions call for three kinds of study: RQ1 is a benchmark comparison that validates the generated data externally, RQ2 is an exploratory characterization of the behavior and interaction that emerge within one run, and RQ3 is an application study of the generated cases on a production assistant.
We first catalog the shared study objects and then describe the data, metrics, and procedure of each study in turn.

\subsection{Study Objects}
The evaluation builds on the artifacts of one simulation run and on one external dataset.
The run was configured with 16 personas in a shared Beijing world and produced two recorded products: an ordered event trace of 4{,}036 committed events, which feeds RQ2, and a sequence of 611 full-system snapshots.
The persona brains are LLM-based (Doubao 2.0 Lite, temperature 0).
Every brain response is recorded in the event trace, so the archived run can be re-audited without re-invoking the model.
Read-only projections derive the remaining two generated objects from the terminal snapshot without modifying the source state.
For RQ1, the projection exports 336 structured personal-information records in six categories: contacts, SMS, call history, alarms, notes, and schedules.
For RQ3, it produces 717 evaluation cases.
The cases span the exported phones of the 16 personas (25 to 57 cases per persona).
Each case carries a snapshot pointer as its source of correctness.
The external object is the held-out real-user benchmark used by RQ1.
It is the production assistant's existing evaluation set, built from real-user data whose provenance is subject to confidentiality and cannot be disclosed, and it is itself never released.
Of its on-device records, 1,186 fall into the same six categories and enter the comparison.
We call these six categories the \textit{shared categories}.

\subsection{RQ1: Benchmark Comparison}
We compare the distributions of the generated data and the benchmark over the shared categories.
We normalize the record counts of the generated data and the benchmark and report Jensen--Shannon divergence (JSD, base-2) and total-variation distance (TVD) over the shared categories.
We further compare content form and temporal rhythm.
For content form, we run two-sample Kolmogorov--Smirnov (KS) tests on the per-category record-length distribution ($\alpha=0.05$).
For temporal rhythm, we compute the JSD of the hour-of-day distribution of record timestamps.
We interpret these as similarity in category structure, content form, and daily rhythm, not as a match to natural usage frequency.
The benchmark is strictly held out: it is never used to generate scenarios, assign labels, or tune persona behavior, so the comparison is an external validation rather than a construction target~\cite{Xu2024BenchmarkLeakageLLMs}.

\subsection{RQ2: Emergence Characterization}
We build a directed interaction graph from the person-to-person dialogue events in the event trace and report node and edge counts, graph density, and reciprocity.
We measure behavioral diversity as the Shannon entropy of each persona's action-type distribution.
No interaction is scripted: the run configuration contains no interaction schedule, and peer interaction arises only from shared-world dynamics, while external inbound messages remain configuration-provided (Section~\ref{sec:scope}).

\subsection{RQ3: Defect Exposure}
The 717 cases are constructed over the exported phone databases of the 16 personas, following the production assistant's existing evaluation scheme.
The query-type and difficulty annotations of each case come with the scheme.
Case labels are assigned by a deterministic labeling component that reads objective world and phone state only and never accesses the benchmark.
The 717 cases are run against the target production assistant in its dedicated evaluation environment.
Each case is annotated with one of five query types (direct entity reference, relational appellation, weak-semantics colloquial, event-topic, time-constrained) and one of three difficulty levels (easy, medium, hard).
An offline checker then evaluates correctness by comparing the assistant's selected record against the stored pointer in the evaluation case, bypassing any LLM judge~\cite{Gu2026LLMasaJudgeSurvey}.
A failure is a case whose selected record does not match the pointer.
We report failure counts and rates by query type and difficulty and diagnose each failure by tracing the pointer back to the snapshot state.

\section{Results}
In this section, we present the evaluation results to answer the three research questions formulated in Section~IV.

\subsection{Answer to RQ1}
RQ1 investigates how closely the distribution of the generated data matches that of the held-out real-user benchmark.
To answer this question, we evaluate the generated data from three complementary perspectives: category structure, measured by JSD and TVD; daily rhythm, measured by the JSD of hour-of-day timestamps; and content form, measured by KS tests on record lengths.

\begin{table}[htbp]
\caption{Category Distribution of the Generated Data and the Held-Out Real-User Benchmark.}
\begin{center}
\begin{tabular}{@{}lcc@{}}
\toprule
\textbf{Category} & \textbf{Gen} & \textbf{Real} \\
\midrule
contacts & 61 (18.1\%) & 320 (27.0\%) \\
SMS & 93 (27.7\%) & 378 (31.9\%) \\
call history & 54 (16.1\%) & 224 (18.9\%) \\
schedules & 51 (15.2\%) & 213 (18.0\%) \\
notes & 50 (14.9\%) & 11 (0.9\%) \\
alarms & 27 (8.0\%) & 40 (3.4\%) \\
\bottomrule
\end{tabular}
\label{tab:rq1}
\end{center}
\end{table}

The first perspective is category structure.
Table~\ref{tab:rq1} lists the generated and benchmark records over the shared categories.
The two distributions are close overall (JSD 0.070, TVD 0.186).
Four categories agree point-wise within five percentage points: SMS (27.7\% vs.\ 31.9\%), call history (16.1\% vs.\ 18.9\%), schedules (15.2\% vs.\ 18.0\%), and alarms (8.0\% vs.\ 3.4\%).
The residual difference concentrates in two categories with opposite directions.
The benchmark holds a larger contact share (27.0\% vs.\ 18.1\%), which we read mainly as an accumulation effect: contact lists on real phones are a stock built up over years, whereas a day-long run reproduces the flow mix but not the accumulated stock.
The generated data instead carries a much larger note share (14.9\% vs.\ 0.9\%). Personas author notes actively during the day, whereas the benchmark records almost none.

The second perspective is daily rhythm.
The hour-of-day distribution of generated timestamps closely matches the benchmark for the two communication categories (JSD 0.064 for SMS and 0.085 for call history).
The match is weaker for schedules (JSD 0.252), alarms (JSD 0.459), and notes (JSD 0.603).
The benchmark histograms point to recording artifacts. Specifically, 34\% of its schedule timestamps fall at 22:00 and 20\% of its alarm timestamps at 00:00, suggesting batch entry rather than natural setting times.
The generated timestamps instead spread across active hours, with 70\% of generated alarms between 6:00 and 9:00.
The notes comparison rests on only 11 benchmark timestamps and is therefore high-variance.
Contacts carry no generated-side timestamps and are excluded from this comparison.

The third perspective is content form.
Table~\ref{tab:rq1len} reports two-sample KS tests on the per-category record-length distributions.
A test rejects a shared length distribution when $D$ exceeds the critical value determined by the two sample sizes at $\alpha=0.05$.
The tests reject for all six categories, but the gap varies widely.
Schedules are nearly identical in length (generated mean 70.7 vs.\ real 75.6 characters, $D=0.234$ barely above its critical value 0.212).
Communication records show the largest substantive gap. Real SMS messages are about twice as long as generated ones (means 114.1 vs.\ 53.1 characters), because they may carry more intents, whereas generated records are single concise entries.
Alarms are the only category where generated records are longer than real ones (means 55.1 vs.\ 44.2 characters).
The note test yields the largest $D$ (0.880) but rests on only 11 benchmark records, so we consider it inconclusive.
We report this content-form gap openly and leave content enrichment to future work.

\begin{table}[htbp]
\caption{Two-Sample KS Tests on Per-Category Record Lengths.}
\begin{center}
\begin{tabular}{@{}lccc@{}}
\toprule
\textbf{Category} & \textbf{Gen mean} & \textbf{Real mean} & \textbf{KS $D$} \\
\midrule
contacts & 22.9 & 44.8 & 0.746 \\
SMS & 53.1 & 114.1 & 0.806 \\
call history & 20.3 & 72.9 & 0.594 \\
schedules & 70.7 & 75.6 & 0.234 \\
notes & 51.6 & 340.0 & 0.880 \\
alarms & 55.1 & 44.2 & 0.527 \\
\bottomrule
\end{tabular}
\label{tab:rq1len}
\end{center}
\end{table}

\AnswertoRQ{RQ1}{Overall, the generated data closely matches the benchmark in category distribution and in the daily rhythm of communication records, though generated records remain shorter and note-taking is more frequent than in the benchmark.}

\subsection{Answer to RQ2}
Having established how closely the generated data matches real distributions, we next examine what the simulation itself produces.
RQ2 investigates what behavior and interaction emerge when personas share one world without scripted interaction.
To answer this question, we evaluate the reported run from three complementary perspectives: interaction structure, measured by the dialogue graph; behavioral diversity, measured by per-persona action-type entropy; and generated content, examined qualitatively.

\begin{figure}[htbp]
\centering
\includegraphics[width=0.7\columnwidth]{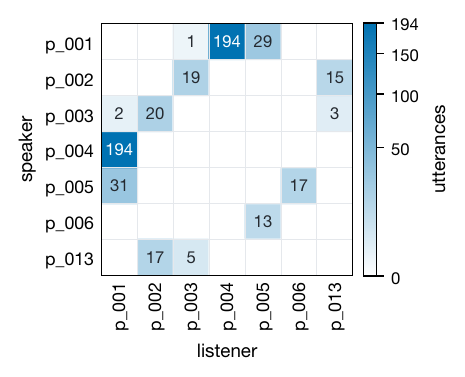}
\caption{Person-to-person dialogue of the active personas.}
\label{fig:dialogue}
\end{figure}

The first perspective is interaction structure.
Fig.~\ref{fig:dialogue} shows the dialogue matrix of the 560 person-to-person dialogue events, restricted to the seven active personas (the nine personas without dialogue are omitted).
The dialogue is concentrated.
Only 7 of the 16 personas engage, and the heaviest pair alone (the spouses \texttt{p\_001} and \texttt{p\_004}) carries 69\% of all utterances.
It is also mutual: the 14 directed edges form 7 fully reciprocated pairs (density 0.333, reciprocity 1.0 within the active subgraph).
No interaction schedule is scripted.
This structure arises from personas meeting and relating inside the shared world.

\begin{figure}[htbp]
\centering
\includegraphics[width=0.8\columnwidth]{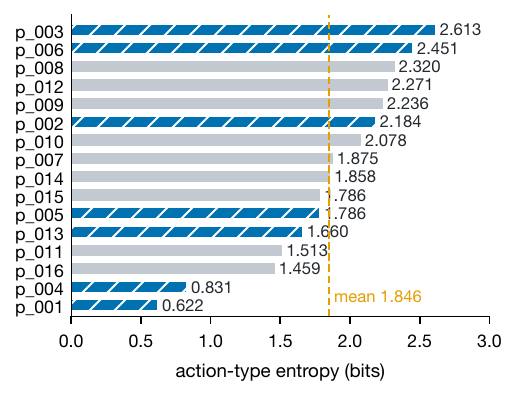}
\caption{Per-persona action-type entropy (bits).
Dialogue-active personas are shown as hatched blue bars.}
\label{fig:entropy}
\end{figure}

The second perspective is behavioral diversity.
Fig.~\ref{fig:entropy} sorts the per-persona action-type entropy, which averages 1.846 bits and spans 0.622 to 2.613 bits.
The two lowest entropies belong to the spouse pair: \texttt{p\_001} (0.622) and \texttt{p\_004} (0.831), whose days are dominated by reflection and each other.
Reflection counts also spread widely, from 33 to 244 across personas.
Repertoires thus differ from persona to persona rather than replaying one template.

The third perspective is generated content.
The run also leaves 1{,}377 thought traces and 560 utterances whose content couples persona identity with world state (translated from the generated traces).
For example, a night-shift nurse (\texttt{p\_014}, 22:52) thinks ``the ward is finally quiet\ldots I will prepare the IV refill for bed 18 before midnight.''
Such excerpts indicate that the agents maintain their personas and situational awareness, a baseline content consistency distinct from the system-level emergence discussed above.

\AnswertoRQ{RQ2}{Without any scripted interaction, personas form a fully reciprocated dialogue subgraph, their behavioral repertoires differ measurably, and their generated content successfully maintains situational awareness.}

\subsection{Answer to RQ3}
Having established what the run produces and how closely it matches real distributions, we finally ask whether the generated data is useful in practice.
RQ3 investigates whether the generated evaluation cases expose and help diagnose defects in a production smartphone assistant.
To answer this question, we evaluate the 717 cases from three complementary perspectives: failure distribution, measured by failure counts and rates across query types and difficulties; defect pattern, measured by the concentration of failures across data types; and defect diagnosis, examined through pointer-based diagnosability and the surface form of the failed queries.

\begin{table}[htbp]
\caption{Production-Assistant Failures by Query Type and Difficulty.}
\begin{center}
\begin{tabular}{@{}lccc@{}}
\toprule
\textbf{Query type} & \textbf{Cases} & \textbf{Failures} & \textbf{Rate} \\
\midrule
direct entity reference & 320 & 4 & 1.3\% \\
relational appellation & 75 & 3 & 4.0\% \\
weak-semantics colloquial & 99 & 1 & 1.0\% \\
event-topic & 150 & 0 & 0.0\% \\
time-constrained & 73 & 0 & 0.0\% \\
\toprule
\textbf{Difficulty} & & & \\
\midrule
easy & 354 & 6 & 1.7\% \\
medium & 264 & 1 & 0.4\% \\
hard & 99 & 1 & 1.0\% \\
\midrule
\textbf{Overall} & \textbf{717} & \textbf{8} & \textbf{1.1\%} \\
\bottomrule
\end{tabular}
\label{tab:rq3}
\end{center}
\end{table}

The first perspective is failure distribution.
Table~\ref{tab:rq3} reports the failure counts and rates categorized by query type and difficulty.
Overall, the assistant answers 709 of the 717 cases correctly (98.9\%).
The high pass rate indicates that the generated cases are processable by a production assistant, so the case set functions as a working evaluation asset end to end.
Notably, the assistant already passes the held-out real-user benchmark; the eight failures arise only over the generated states, so they are defects that the existing evaluation set does not detect.
Across query types, relational appellation is the hardest (4.0\%), since answering it requires first resolving a social-role appellation (such as ``my wife'') to a contact entity, whereas event-topic and time-constrained queries never fail.
The failure rate does not increase with difficulty (easy 1.7\%, medium 0.4\%, hard 1.0\%), so the annotated difficulty does not predict where the assistant fails.

The second perspective is the defect pattern.
All eight failures (100\%) fall on communication records: 5 of 77 call-history cases (6.5\%) and 3 of 209 SMS cases (1.4\%) fail, whereas contacts, schedules, notes, and alarms never fail.
The generated cases thus separate where the assistant is reliable from where it is not, although the zero-failure categories may partly reflect less challenging cases rather than mastered behavior.

The third perspective is defect diagnosis.
Because the pointer denotes a record that exists in the exported state by construction, it isolates assistant-side errors from data-generation errors, so all eight failures are diagnosable without an LLM judge.
Inspecting the failed queries shows a consistent surface form: the failed queries predominantly anchor the target record through a message direction (such as a message the user sent versus received) or a relational appellation (such as one's spouse or son), and the two anchors frequently co-occur.
This case-level pattern agrees with the set-level one, in which relational appellation is the hardest query type and every failure falls on communication records.

\AnswertoRQ{RQ3}{The 717 evaluation cases expose eight failures in the production assistant, five on call history and three on SMS. Checking and diagnosis require no LLM judge.}

\section{Conclusion and Future Work}
We presented SenWorld, a physically grounded, deterministic, event-sourced digital-twin simulation that advances personas, phones, services, and physical contexts as one evolving world and archives a full-system snapshot at every frame.
Because the simulation owns its state and evolves it under real physical constraints, whatever is true at a simulated moment is itself labeled data: an evaluation case is answered by a snapshot pointer to an existing record, and checking the answer requires no LLM judge.
Three studies evaluated the generated data.
The generated data closely matches the held-out real-user benchmark in category distribution (JSD 0.070) and in the daily rhythm of communication records (JSD below 0.1), while the gaps in record length and note-taking frequency remain open.
Without any scripted interaction, personas form a fully reciprocated dialogue subgraph and measurably different behavioral repertoires.
Projected into 717 evaluation cases, the generated data runs against a production smartphone assistant without an LLM judge: the assistant answers 98.9\% of the cases, and all eight residual failures fall on call and SMS records, where each remains diagnosable through its snapshot pointer.

Two directions remain open.
First, a controlled ablation would isolate the causal contribution of physical grounding to fidelity, which this paper attributes by design rather than by experiment.
Second, grounding the world in more cities, across seeds, and over longer simulated horizons would test whether the reported fidelity and emergence properties hold at scale.
The benchmark, the generated data, and the evaluation cases cannot be released under the applicable data-governance policy.
Overall, SenWorld offers a privacy-safe, reproducible, and distribution-checked path to evaluation data whose ground truth is fixed by construction.

\bibliographystyle{IEEEtran}
\bibliography{references}

\end{document}